\documentclass{article}

\usepackage{microtype}
\usepackage{graphicx}
\usepackage{subfigure}
\usepackage{booktabs} 
\usepackage{amsmath}
\usepackage{amsfonts}
\usepackage{bm}
\usepackage{hyperref}

\usepackage[accepted]{icml2019}

\icmltitlerunning{Learning Accurate Extended-Horizon Predictions of High Dimensional Trajectories}

\begin{document}

\twocolumn[
\icmltitle{Learning Accurate Extended-Horizon Predictions of High Dimensional Trajectories}


\icmlsetsymbol{equal}{*}

\begin{icmlauthorlist}
\icmlauthor{Brian Gaudet}{deep}
\icmlauthor{Richard Linares}{mit}
\icmlauthor{Roberto Furfaro}{uofa}
\end{icmlauthorlist}

\icmlaffiliation{mit}{Charles Stark Draper Assistant Professor, Department of Aeronautics and Astronautics, Massachusetts Institute of Technology}
\icmlaffiliation{deep}{Brian Gaudet, Co-Founder, DeepAnalytX LLC, University of Arizona Department of Systems and Industrial Engineering, Department of Aerospace and Mechanical Engineering}
\icmlaffiliation{uofa}{Professor, University of Arizona Department of Systems and Industrial Engineering, Department of Aerospace and Mechanical Engineering}
\icmlcorrespondingauthor{Brian Gaudet}{briangaudet@mac.com}

\icmlkeywords{Reinforcement Learning Learning, Model based planning}

\vskip 0.3in
]



\printAffiliationsAndNotice{}  

\begin{abstract}
We present a novel predictive model architecture based on the principles of predictive coding that enables open loop prediction of future observations over extended horizons. There are two key innovations.  First, whereas current methods typically learn to make long-horizon open-loop predictions using a multi-step cost function, we instead run the model open loop in the forward pass during training.  Second, current predictive coding models initialize the representation layer's hidden state to a constant value at the start of an episode, and consequently typically require multiple steps of interaction with the environment before the model begins to produce accurate predictions. Instead, we learn a mapping from the first observation in an episode to the hidden state, allowing the trained model to immediately produce accurate predictions.  We compare the performance of our architecture to a standard predictive coding model, and demonstrate the ability of the model to make accurate long horizon open loop predictions of simulated Doppler radar altimeter readings during a six degree of freedom Mars landing. Finally, we demonstrate a 2X reduction in sample complexity by using the model to implement a Dyna style algorithm to accelerate policy learning with proximal policy optimization.
\end{abstract}

\section{Introduction}
\label{introduction}

A model capable of accurate multi-step prediction prediction over long horizons has the potential to reduce the sample complexity of reinforcement learning as compared to model free methods.  For example, such a model would allow an agent to plan by estimating the effect of some sequence of actions.  Alternately, the model can be used in a Dyna \cite{sutton1990integrated} style algorithm to generate rollouts starting from observations sampled from a replay buffer. A fundamental trade-off exists when learning a model suitable for model-based reinforcement learning. Complex dynamics necessitate a complex model, which is prone to overfitting without large amounts of data.  However, for model-based RL, we want the model to learn a good representation of the dynamics quickly to reduce sample complexity.  If learning the model is as difficult as learning a model-free policy, we have gained nothing (unless the model can be re-used for other tasks).

In this work we develop a predictive model using the principles of predictive coding \cite{rao1999predictive}, which was originally developed to explain endstopping in receptive fields of the visual cortex. An action conditional predictive coding model maps from a history of prediction errors and actions to the predicted observation $f: \mathbf{e}_{0:t}, \mathbf{a}_{0:t} \mapsto \hat{\mathbf{o}}_{t+1}$, where $\mathbf{e}_{t+1}=\hat{\mathbf{o}}_{t+1} - \mathbf{o}_{t+1}$, and $\mathbf{e}_0=\mathbf{0}$. When the trained model is run open loop to make predictions, the error feedback signal $\mathbf{e}$ is set to zero, i.e., the model implements the function $f: \mathbf{0}, \mathbf{a}_{0:t} \mapsto \hat{\mathbf{\hat{o}}}_{t+1}$, which is equivalent to using the predicted observation in place of the measured observation to compute the error feedback.

In contrast, most action conditional models reported in the literature implement a mapping from a history of observations and actions to a predicted observation $f: \mathbf{o}_{0:t}, \mathbf{a}_{0:t} \mapsto \hat{\mathbf{o}}_{t+1}$.  The model can then be used to predict a trajectory open loop given some initial observation and a sequence of actions by feeding back the predicted observation into the model $f: \mathbf{\hat{o}}_{0:t}, \mathbf{a}_{0:t} \mapsto \hat{\mathbf{o}}_{t+1}$, where in model-based RL, the action would be output by some policy $\pi: \mathbf{o} \mapsto \mathbf{a}$. 

Our model borrows from the PredNet architecture \cite{lotter2016deep}, which applies modern deep learning techniques to predictive coding, but is modified to make the model action conditional, and implements two key innovations. To understand the first improvement, first consider that a predictive coding model (PCM) does not have access to the observations, but only to the prediction error.  Consequently, it typically takes several steps of closed loop interaction with the environment for a trained PCM to begin making accurate open loop predictions, which can be a problem in some applications. Our solution is to learn a mapping from the initial observation in an episode to the model's recurrent network hidden state $f: \mathbf{o_0} \mapsto \mathbf{h_0}$.  This  effectively bootstraps the hidden state, allowing the model to immediately begin making accurate predictions from the onset of an episode. Concretely, the model implements a mapping $f: \mathbf{o}_0, \mathbf{e}_{0:t}, \mathbf{a}_{0:t} \mapsto \hat{\mathbf{o}}_{t+1}$. 

The second improvement concerns the forward pass through the model's network during training. If the network is unrolled for T steps to implement back propagation through time, we input the prediction error $\mathbf{e}$ only on the first step, and set $\mathbf{e}=\mathbf{0}$ for steps 2 through T. This forces the network to learn a representation that is good for multi-step prediction, even when using a conventional single step cost function. Open loop prediction of trajectories is  accomplished by initializing the recurrent network's hidden state using the learned mapping from the initial observation at the start of the episode, and setting the prediction error $\mathbf{e}$ to zero, i.e., the trajectory is generated using  the mapping $f: \mathbf{o}_0, \mathbf{0}, \mathbf{a}_{0:t} \mapsto \hat{\mathbf{o}}_{t+1}$. 

We demonstrate through a series of experiments that even without our enhancements, action conditional predictive coding allows accurate open-loop trajectory prediction of high dimensional trajectories \footnote{Note that here we are referring to the dimension of the configuration space that the agent is operating in, not the dimension of the measurement space. For example, an agent operating in a simulated Atari environment has a low dimensional configuration space but a high dimension measurement space} over long horizons, and that our refinements further enhance this accuracy. We attribute the high performance of predictive coding to the fact that a predictive coding model must make predictions using only a history of prediction errors and actions (as opposed to observations and actions). Since the model must integrate a history of prediction errors, minimizing the loss requires the model to learn important temporal dependencies.

In these experiments we use two versions of a  Mars powered descent phase environment. The first version is a simple 3 degree of freedom (3-DOF) environment that is used to compare a standard PCM to a model with our enhancements.  Although only 3-DOF, the environment is realistic with respect to the lander capability, dynamics, and deployment ellipse. The model is then tested in a 6-DOF version of the Mars landing environment.  In one experiment we demonstrate  the model's ability to predict the lander's $\mathbb{R}^{13}$ trajectory open loop over an entire episode, whereas in the second experiment the model must predict simulated Doppler radar altimeter readings open loop during the descent, with the simulated readings generated using a digital terrain map of the Uzboi Vallis region on Mars. We then couple the model with an agent implementing proximal policy optimization (PPO) \cite{schulman2017proximal} and demonstrate a reduction in sample complexity in the 3-DOF  landing task using a Dyna style algorithm. 

To our knowledge, this is the first demonstration of a sample efficiency improvement of PPO using concurrent training of policy and model, as previous work has focused on accelerating Q learning. Although \cite{nagabandi2018neural} used TRPO, the training had two distinct phases of model-based and model free learning, with model predictive control used in the 1st phase. In addition, most previous work in model-based acceleration has assumed that a ground-truth reward function is known, e.g., see \cite{nagabandi2018neural}, \cite{gu2016continuous}, and \cite{feinberg2018model}. In contrast, our model learns a reward function using a separate reward head in the model network.

\section{Preliminaries}

Our goal is for the model to learn a representation that is effective for making long-term open loop predictions of high dimensional trajectories. In this work the model will learn to predict future observations in a reinforcement learning setting where an agent interacts with an environment and learns to accomplish some task, with the learning driven by a scalar reward signal. The agent will instantiate a policy $\pi: \mathbf{o} \mapsto \mathbf{a}$ that maps observations to actions, and the model learns a representation by observing the sequence of observations and actions induced by the episodic interaction between the agent and environment. Ideally,  after a limited amount of experience observing this interaction, the trained model will have the ability to accurately predict future observations from some sequence of on-policy actions, operating open loop. This ability to "imagine" the long-term consequences of some sequence of actions will then be a powerful tool for planning in model-based reinforcement learning, or alternatively, could be used to accelerate model-free algorithms by augmenting the rollouts collected via interaction with the environment with simulated rollouts from the model. 

We will evaluate our model using two criteria.  The first is the ability of the model to be able to generate accurate extended horizon trajectory predictions given some sequence of actions, with the policy that is generating actions having access to the ground truth observation. The second criterion is whether the model can significantly accelerate proximal policy optimization (PPO), a model-free policy gradient with baseline algorithm. 


\section{Related Work}

Recent work in developing predictive models include \cite{finn2017deep}, where a model learns to predict future video frames by observing sequences of observation and actions.  The model is then used to generate robotic trajectories using model predictive control (MPC), choosing the trajectory that ends with an image that best matches a user specified goal image. The action conditional architecture of \cite{oh2015action} has proven successful in open loop prediction of long sequences of rendered frames from simulated Atari games. Predictive coding \cite{lotter2016deep} has been been applied to predicting images from objects that are sequentially rotated, and has been used to predict steering angles from video frames captured from a car mounted camera. \cite{wang2017predrnn} and \cite{wang2017predrnn} extend the PredNet architecture described in \cite{lotter2016deep}.

As for using predictive models to accelerate reinforcement learning, in \cite{nagabandi2018neural}, the authors use an action conditional model to predict the future states of agents operating in various openAI gym environments with high dimensional state spaces, and use the model in a model predictive control algorithm that quickly learns the tasks at a relatively low level of performance.  The model is then used to create a dataset of trajectories to pre-train a TRPO policy \cite{schulman2015trust}, which then achieves a high level of performance on the task through continued model-free policy optimization. The authors report a 3 to 5 times reduction in sample efficiency using the combined model free and model based algorithm. The approach assumes that a ground truth reward function is known. 

In \cite{gu2016continuous}, the authors develop a normalized advantage function that gives rise to a Q-learning architecture suitable for continuous high dimensional action spaces.  They then improve the algorithm's sample efficiency by adding "imagination rollouts" to the replay buffer, which are created using a time varying linear model. Importantly, the authors state that their architecture had no success using neural network based models to improve sample efficiency, and assume a ground-truth reward function is available. 

\cite{feinberg2018model} accelerates learning a state-action value function by using a neural network based model to generate synthetic rollouts up to a horizon of H steps.  The rewards beyond H steps are replaced by the current estimate of the value of the penultimate observation in the synthetic rollouts, as in an H-step temporal difference estimate. The accelerated Q function learning is then used to improve the sample efficiency of the DDPG algorithm \cite{lillicrap2015continuous}. Again, the algorithm assumes that the ground-truth reward function is available.

\section{Learning to predict over long horizons}

Although a model can easily learn a representation mapping observations and actions at step $t$ to a predicted next observation at step $t+1$ (one-step predictions), these representations typically do not capture the underlying dynamics of the system.  The equations of motion encoding the laws of physics typically constrain states in configuration space that are close to each other in time to be numerically close in value, and the observations that are functions of these states (such as a sequence of video frames) will by similar for two consecutive frames.  Consequently, the model's network will tend to learn a trivial mapping that does not capture the underlying dynamics.  When such a network is run forward open loop to make multi-step predictions (as required when using such a model for planning), the prediction accuracy falls of rapidly as a function of the number of steps. 

\subsection{Using Differences in Observations as Training Targets}
Some environments (such as many of the open AI Mujoco environments) use dynamical systems where the system dynamics are dominated by rigid body rotations. For example, the half-cheetah environment has an observation space $S \in \mathbb{R}^{17}$, but the translational motion is constrained to $\mathbb{R}^2$ (position and velocity along a line).  In these dynamical systems,  having an action conditional model's network use the difference between consecutive observations (deltas) as a training target can lead to a representation with improved multi-step prediction performance (see for example \cite{nagabandi2018neural}). Concretely, the training target is given as $o_{t+1}-o_{t}$ for all $t$ in the rollouts.      

In our work applying RL to aerospace problems, we have found that using deltas as targets fails for environments  where translational motion plays a more important roll in the system dynamics. We postulate that the reason using deltas as targets works well for rotational dynamics is that the mapping between torques and rotational velocities using Euler's rotational equations of motion has sufficient complexity to insure that the model does not learn an identity mapping between components of the current observation and components of the deltas. On the other hand,  the differential equations governing position,  velocity, and commanded thrust are such that using deltas actually makes matters worse, as the delta velocity component of the training target is approximately proportional to the force (the action input) and the delta position component of the training target is proportional to the velocity input to the model.



\subsection{k-step prediction loss functions}

Another common approach to improving multi-step prediction is to use a K-step prediction loss function \cite{oh2015action}.  Consider training data accumulated over $M$ episodes, with $T$ steps in a given episode. Then the K-step loss function is as shown below in Equation \ref{eq:kstep_loss}, with the more commonly used 1-step loss given in \ref{eq:1step_loss}. Minimizing the K-step loss function requires the model to learn a representation that is good at multi-step predictions.  However, this approach also increases the size of the data used for a model update by a factor of K, as each sample from the rollouts is used to produce a K-step open loop prediction. Moreover, \cite{oh2015action} required curriculum learning \cite{bengio2009curriculum}, with the value of K was increased once a given value of K converged.

\begin{subequations}
    \begin{align}
	L(\theta)=\frac{1}{2MTK}\sum_i^M\sum_t^T\sum_{k=1}^K\|\hat{\mathbf{o}}_i^{(t+k)}-\mathbf{o}_i^{(t+k)}\|\label{eq:kstep_loss}\\
	L(\theta)=\frac{1}{2MT}\sum_i^M\sum_t^T\|\hat{\mathbf{o}}^{(i)}-\mathbf{o}^{(i)}\|\label{eq:1step_loss}
	\end{align}
\end{subequations}

\subsection{Open Loop Forward Pass}

Recently, we have found a new method to improve multi-step prediction performance in predictive models that use one or more recurrent layers. First, we will review a common approach to implementing the forward pass in a network with one or more recurrent layers that allows parallel computation over a sequence where we want to preserve the temporal dependencies. Prior to the recurrent layer(s), the  network unrolls the output of the previous layer, reshaping the data from $\mathbb{R}^{m\times n}$ to $\mathbb{R}^{T\times (m/T)\times n}$, where $m$ is the batch size, $n$ the feature dimension, and $T$ the number of steps we unroll the network. We then implement a loop from step 1 to step $T$, where at step one, we input the hidden state from the rollouts to the recurrent layer, but for all subsequent steps up to $T$, we let the state evolve according to the current parameterization of the recurrent layer. After the recurrent layer (or layers), the recurrent layer output is then reshaped back to $\mathbb{R}^{m\times n}$.

Now consider a modification to this approach where we unroll the entire network from the network's inputs to the final recurrent layer, as opposed to just unrolling the recurrent layers.  For this example, we will assume a predictive coding architecture. At the first step of the loop over the T steps the network is unrolled, we input the prediction error $\mathbf{e}$ and action $\mathbf{a}$ from the rollouts to the network's first layer. However, on subsequent timesteps, we set $\mathbf{e}$ to zero, making the remainder of the unrolling open loop, in that there is no feedback from the prediction error captured in the rollouts. Consequently, to minimize the 1-step cost function, the network must learn a representation that is useful for multi-step prediction. This approach is also applicable to action conditional models.  Here we need to unroll the entire network so that we can feed back the predicted observation as the observation for steps 2 through T.

This approach has the advantage over a k-step  prediction loss function in that it does not increase the training set size by a factor of K.  Moreover, in contrast to \cite{oh2015action}, this method does not require curriculum learning. 

\section{Predictive Coding Model used in Experiments}
\label{section:pcm_exp}
Our model's operation over a single episode is listed in Algorithm \ref{alg:pcm}, where $\pi: \mathbf{o} \mapsto \mathbf{a}$ is a policy, $\mu: \mathbf{e},\mathbf{a} \mapsto \mathbf{\mathbf{\hat{o}}}, v$ is our model, with both operating in environment $env$.  The model  consists of four major components.  A  representation network  maps the previous prediction error $\mathbf{e}$  and current action $\mathbf{a}$ to a representation $\mathbf{r}$ (Eq. \ref{eq:pcm_r}), and consists of a fully connected layer followed by a recurrent layer implemented as a gated recurrent unit (GRU)\cite{chung2015gated} with hidden state $\mathbf{h}$.  

A  prediction network consists of two fully connected layers, and maps the representation $\mathbf{r}$ to a predicted observation $\mathbf{\hat{o}}$ (Eq. \ref{eq:pcm_p}).  In an application using high dimensional observations, the prediction network would contain several deconvolutional layers mapping from the representation to a predicted image $\mathbf{\hat{o}}$. 

The recurrent state initialization network learns a mapping from an observation at the start of an episode $\mathbf{o}_0$ to an initialization for hidden state $\mathbf{h}$ in the representation network $\mathbf{h}$ (Eq. \ref{eq:pcm_i}). This allows the trained model to immediately produce accurate predictions.

Finally the value prediction network maps the representation $\mathbf{r}$ to $v$, an estimate of the sum of discounted rewards that would be received by starting in representation $\mathbf{r}$ and following policy $\pi$ (Eq. \ref{eq:pcm_v}).  Since in theory a value function could be implemented in an external network using representation $\mathbf{r}$ as input, the  primary purpose of the value network is to solve the "vanishing bullet" problem when the model is used with visual observations. The vanishing bullet problem, coined for the Atari space invaders environment, occurs when a very important visual feature consists of relatively few pixels. In this case, a model using an L2 loss between predicted and actual observations might not include the "bullets" fired by the space invaders in its predictions, leading to rather poor performance if the model were to be used for planning. 

Note that the value prediction head can be replaced by a reward prediction head by simply changing the target from the sum of discounted rewards to rewards.  In section \ref{section:Dyna}, we use the reward prediction head in our experiment where we use the model to accelerate learning. In applications using images as observations, predicting rewards should have a similar effect on mitigating the "vanishing bullet" problem, except possibly in sparse reward settings. The network layers are detailed in Figure \ref{tab:network}, where $d_o$ is the observation dimension and $d_{o+a}$ is the sum of the observation and action dimensions.

Our model is trained end-to-end using two L2 loss functions, one for the difference between the predicted and actual observation and another for the difference between the predicted value and the empirical sum of discounted rewards. Similar to the model used in \cite{lotter2016deep}, this model can be treated as a layer and stacked, providing higher levels of abstraction for higher layers.  This stacking can enhance performance when observations consist of images, but we found it unnecessary in this work.

\begin{subequations}
\begin{align}
	\mathcal{R}&: \mathbf{e}, \mathbf{a} \mapsto \mathbf{r} \label{eq:pcm_r}\\
	\mathcal{P}&: \mathbf{r} \mapsto \mathbf{\hat{o}}\label{eq:pcm_p}\\
	\mathcal{I}&: \mathbf{o}_0 \mapsto \mathbf{h}\label{eq:pcm_i}\\
	\mathcal{V}&: \mathbf{r} \mapsto \hat{v}\label{eq:pcm_v}
\end{align}
\end{subequations}

\begin{table}[H]
\caption{Network Layers}
\label{tab:network}
\vskip 0.15in
\begin{center}
\begin{small}
\begin{sc}
\begin{tabular}{lcccc}
\toprule
\midrule
Layer & in dim & out dim & type & act \\
$\mathcal{R}_1$ & $d_o $ & $10\times d_{o+a} $ & FC & tanh\\
$\mathcal{R}_2$ & $10\times d_{o+a}$ & $10\times d_{o+a} $ & GRU & None\\
$\mathcal{P}_1$ & $10\times d_{o+a} $ & $10\times d_{o+a} $ & FC & tanh\\
$\mathcal{P}_2$ & $10\times d_{o+a} $ & $d_o $ & FC & None\\
$\mathcal{I}_1$ & $d_{o} $ & $10\times d_{o+a} $ & FC & tanh\\
$\mathcal{I}_2$ & $10\times d_{o+a} $ & $d_h $ & FC & tanh\\
$\mathcal{V}_1$ & $10\times d_{o+a} $ & $10\times d_{o+a} $ & FC & tanh\\
$\mathcal{V}_2$ & $10\times d_{o+a} $ & 1  & FC & None\\
\bottomrule
\end{tabular}
\end{sc}
\end{small}
\end{center}
\vskip -0.1in
\end{table}

\begin{algorithm}[tb]
   \caption{PCM Operation}
   \label{alg:pcm}
\begin{algorithmic}
    \STATE Initialize $\mathbf{e}_0=\mathbf{0}$, $t=0$, $\mu.h=\mu(\mathbf{o}_0)$,$\mathbf{o}_0$ = env.reset().
    \WHILE{not env.done}
    \STATE $\mathbf{a}_t=\pi(\mathbf{o}_t)$
    \STATE $\mathbf{o}_{t+1},r=\mathrm{env.step}(\mathbf{a}_t)$
    \STATE $\mathbf{\hat{o}}_{t+1}, \hat{v}_{t+1} = \mu(\mathbf{e}_t,\mathbf{a}_t)$
    \STATE $\mathbf{e}_{t+1}=\mathbf{\hat{o}}_{t+1}-\mathbf{o}_{t+1}$
    \STATE $t=t+1$
    \ENDWHILE
\end{algorithmic}
\end{algorithm}


\section{Experiments}

\subsection{Comparison of standard and enhanced PCM}

Here we compare the performance of a standard predictive coding model and the model described in Section \ref{section:pcm_exp} using a simple 3-DOF Mars landing environment, with initial lander conditions in the target-centered reference frame as shown in Table \ref{tab:IC}. The agent's goal is to achieve a soft pinpoint landing with a terminal glideslope of greater than 5; a complete description of the environment  can be found in \cite{gaudet2018deep}. For this comparison we use a policy implementing the DR/DV guidance law \cite{d1997optimal}, which maps position and velocity to a commanded thrust vector. This is similar to using a pre-trained policy as good trajectories are generated from the onset.

The model is updated using 30 episode rollouts and trained for 30,000 episodes. During the forward pass, the networks are unrolled 60 steps. After training, each model is evaluated on a 3600 episode test where the model is run forward open loop for 1, 10, 30, and 60 steps. This is the same process used to collect rollouts for model and policy training, but we reshape the rollouts to allow running multi-step predictions in parallel on the full set of rollouts, and collect prediction accuracy statistics at each step. Note that in most of these cases, the model state will have evolved since the start of an episode, which is why the average accuracy for the model without hidden state initialization is not that bad. The prediction accuracy is measured as the absolute value of the prediction error (position and velocity) and the value estimate accuracy is measured using explained variance.  Note that model predictions are scaled such that each element of the prediction has zero mean and is divided by three times the standard deviation, and the error is calculated over the samples and features (which have equal scales).  So a prediction error of 0.01 is 1\% of the range of values encountered over all elements of the predicted position and velocity vectors; if the maximum altitude was say 2400m, then a mean absolute value of prediction error of 1\% would correspond to 24m. To insure a fair comparison, the code is identical in each case, except that the appropriate network is instantiated in the model.

\begin{table}[H]
    \caption{Mars Lander Initial Conditions}
   \label{tab:IC}
        \centering 
   \begin{tabular}{l | r | r | r | r } 
      \hline 
       & \multicolumn{2}{c}{Velocity (m/s)}\vline & \multicolumn{2}{c}{Position (m)}\\
       \hline
       & min & max & min  & max  \\
       \hline
      Downrange      & -70 & -10 & 0 & 2000\\
      Crossrange       & -30  & 30 & -1000 & 1000 \\
      Elevation     & -90 & -70 & 2300 & 2400 \\
   \end{tabular}
\end{table}

Tables \ref{tab:model-pred_perf} and \ref{tab:model-ev_perf} shows prediction error performance for three model architectures. An explained variance of '-' indicates less than zero. PCM is a standard predictive coding model, i.e.,  without the mapping between initial observations and the hidden state and without zeroing the error feedback in the forward pass. The maximum prediction error for this model is high (close to the maximum range of each state variable), and occurs when the multi-step error checking begins at the start of an episode, so the model never gets any error feedback.  When the error checking begins mid-episode, performance improves, and consequently the mean error does not look too bad.  PCM-I adds the network layers that learn a mapping between an episodes initial observation and the recurrent layer's hidden state. Here we see that the mean prediction error improves, and although not shown, the maximum prediction error is much reduced.  PCM-I-OLT is our PCM model as described in Section \ref{section:pcm_exp}, where the error feedback is set to zero during the forward pass in training. We see that running the forward pass open loop has a significant impact on multi-step prediction performance. We also include two standard action conditional models as comparison baselines. ACM is a standard action conditional model and ACM-OLT is modified to feed back the predicted observation as the observation during the training forward pass. 

Note that even without the open loop forward pass, the predictive coding model excels at multi-step prediction. We hypothesize that the performance is due to the model taking a prediction error rather than an observation as an input. This requires the model to integrate multiple steps of previous errors in order to make good predictions,  thereby forcing the model to learn a representation capturing temporal dependencies. The model learns reasonably quickly, with the model's performance as a function of training steps given in Table \ref{tab:model-ep_perf}. These statistics are captured during training, and importantly, are measured on the current rollouts before the model trains on these rollouts, and are consequently a good measure of generalization.


\begin{table}[H]
\caption{ K Step Mean Abs Prediction Error}
\label{tab:model-pred_perf}
\vskip 0.15in
\begin{center}
\begin{small}
\begin{sc}
\begin{tabular}{lccccc}
\toprule
\midrule
K           & 1 & 10 & 30 & 60\\
ACM         & 0.0150 & 0.3892  & 0.3605 & 0.3152\\
ACM-OLT     & 0.0165 & 0.1615  & 0.1954 & 0.2174\\
PCM         & 0.0360 & 0.0392  & 0.0439 & 0.0483 \\
PCM-I       & 0.0019 & 0.0088  & 0.0153 & 0.0216 \\
PCM-I-OLT   & 0.0020 & 0.0024  & 0.0033 & 0.0044  \\
\bottomrule
\end{tabular}
\end{sc}
\end{small}
\end{center}
\vskip -0.1in
\end{table}

\begin{table}[H]
\caption{ K Step Mean Explained Variance}
\label{tab:model-ev_perf}
\vskip 0.15in
\begin{center}
\begin{small}
\begin{sc}
\begin{tabular}{lccccc}
\toprule
\midrule
K           & 1 & 10 & 30 & 60\\
ACM         &  0.9380 & - & - & -\\
ACM-OLT     &  0.9759 & 0.6626 & 0.3816 & - \\
PCM         &  0.9808 & 0.9789 & 0.9736 & 0.9604 \\
PCM-I       &  0.9851 & 0.9723 & 0.9426 & 0.9167\\
PCM-I-OLT   &  0.9859 & 0.9877 & 0.9847 & 0.9798 \\
\bottomrule
\end{tabular}
\end{sc}
\end{small}
\end{center}
\vskip -0.1in
\end{table}

\begin{table}[H]
\caption{PCM 20-step Prediction Performance as function of Training Steps}
\label{tab:model-ep_perf}
\vskip 0.15in
\begin{center}
\begin{small}
\begin{sc}
\begin{tabular}{lccccc}
\toprule
\midrule
Steps $\times1e4$ & 8 & 16 & 24 & 78 \\
Pred. Err.     & 0.1155 & 0.0411 & 0.0176 &  0.0065\\
Exp. Var.     & 0.7214  & 0.8536  & 0.8626  & 0.9374 \\
\bottomrule
\end{tabular}
\end{sc}
\end{small}
\end{center}
\vskip -0.1in
\end{table}


\subsection{6-DOF Mars Landing}

Here we test the model's ability to predict a lander's 6-DOF state during a simulated powered descent phase. To make this more interesting, we let the model and agent learn concurrently.  The agent uses PPO to learn a policy to generate a thrust command for each of the lander's four thrusters, which are pointed downwards in the body frame. Rather than use a separate value function baseline, the policy uses the model's value estimate. This is a difficult task, as to achieve a given inertial frame thrust vector, the policy needs to figure out how to properly rotate the lander so that the thrusters are properly oriented, but this also affects the lander's translational motion. The goal is for the lander to achieve a soft pinpoint landing with the velocity vector directed predominately downward, an upright attitude, and negligible rotational velocity. The initial conditions are similar to that given in Table \ref{tab:IC}, except that the lander's attitude and rotational velocity are perturbed from nominal.  A full description of the environment can be found in \cite{gaudet2018deep}.  In the first experiment, the policy and model get access to the ground truth state (position, velocity, quaternion attitude representation, and rotational velocity). 

In a variation on this experiment, the model only gets access to simulated Doppler radar altimeter returns, using a digital terrain map (DTM) of the Mars surface in the vicinity of Uzboi Valis. We assume a configuration similar to the Mars Science Laboratory lander, where there are four altimeters with fixed orientation in the body frame, each pointing downwards and outwards at a 22 degree angle from the body frame vertical axis. Since the DTM spans 40 square km, we had to cut some corners simulating the altimeter readings in order to reduce computation time enough to be practical for the large number of episodes (300,000) required to solve this problem using RL. This results in low accuracy, particularly at lower elevations, further complicating this task. Also, due to the relatively high pitch and roll limits we allow the lander, the altimeter beams occasionally completely miss the DTM, and return a max range reading.

Training updates use 120 episode rollouts. Learning a good policy in 6-DOF typically takes around 300,000 episodes, which is over 80M steps of interaction with the environment. Similarly, we find that model takes longer to learn a good representation.  There are two factors that can contribute to the increased model learning time. First, the dynamics are more complex, and the model's  network is larger.  Second, since it takes a long time for the policy to converge, exploration causes the model to be presented with a wide range of actions for each observation during each training epoch. Table \ref{tab:6-DOF_conv} shows the model's convergence during training as a function of training steps. We look at this for both a pre-trained policy and concurrent learning to establish the primary factor behind the slower model convergence. Although model convergence is slower in the case of concurrent learning, the difference is not extreme; consequently we attribute the slower convergence in the 6-DOF case primarily to the more complicated dynamics. It may be possible to speed model convergence by either increasing the number of model training epochs, training on a larger set of rollouts using a replay buffer, or both.

Table \ref{tab:6-DOF_perf} tabulates the model's performance for both the case where the model has access to the ground truth lander state and the case where the model only has access to simulated Doppler radar returns.  For the latter case, the observation consists of four simulated altitudes and closing velocities, one for each altimeter. These observations do not satisfy the Markov property in that there are many lander states in configuration space that can give identical readings. However, the recurrent network layer allows the model to make reasonably accurate predictions. 

\begin{figure*}[ht]
\vskip 0.2in
\begin{center}
\centerline{\includegraphics[width=\columnwidth*2]{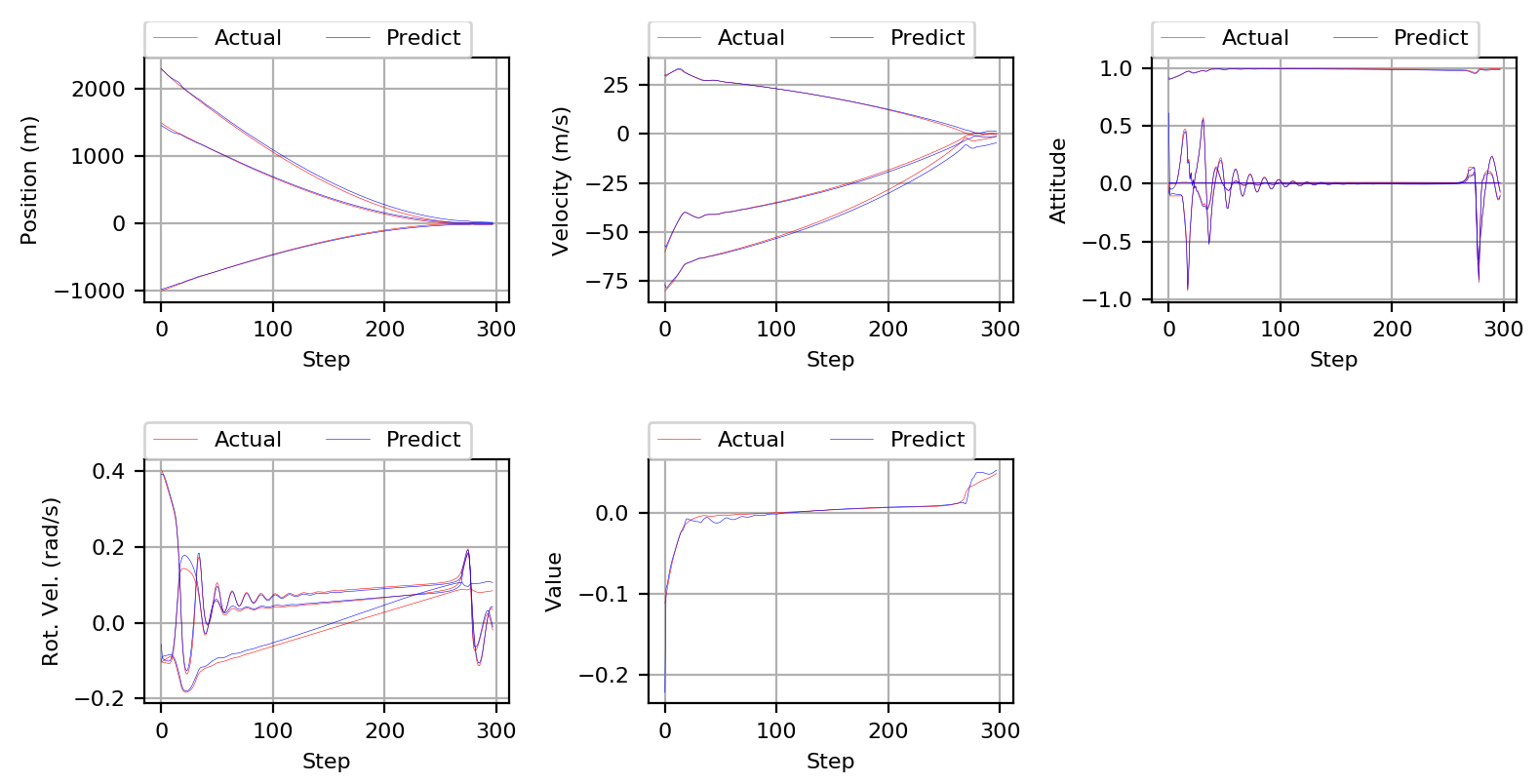}}
\caption{6-DOF Open Loop PCM Prediction over an entire Episode}
\label{fig:PCM_traj_6dof}
\end{center}
\vskip -0.2in
\end{figure*}

Figure \ref{fig:PCM_traj_6dof} illustrates an entire episode of PCM open loop predictions. Here the policy  gets access to the ground truth observation on all steps, but the model only gets access to the ground truth observation on the 1st step of the episode, and for remaining steps, the PCM's prediction error input is set to zero, removing any feedback from actual observations. The "Value" plot shows the estimated value (sum of discounted rewards) of the observation at the current step as predicted by the model's value head. The prediction error is small enough that is is barely discernible from the plot. We use a quaternion representation for attitude. Note that the model's network was unrolled only 60 steps during training, but this sufficed to allow good prediction out to the end of a 300 step episode.

\begin{table}[H]
\caption{ 6-DOF 30 step Mean Abs Prediction Error as function of steps for pre-trained policy (TP) and concurrent learning (LP)}
\label{tab:6-DOF_conv}
\vskip 0.15in
\begin{center}
\begin{small}
\begin{sc}
\begin{tabular}{lccccc}
\toprule
\midrule
Steps $\times1e6$      & 0.37 & 0.74 & 1.11 & 3.69\\
$\mu$ TP      & 0.0883 &  0.0411 & 0.0290   &  0.0155  \\
$\sigma$ TP       & 0.1166 &  0.0649  & 0.0464  &  0.0241\\
$\mu$ LP     & 0.0964 & 0.0575  & 0.0490  & 0.0341   \\
$\sigma$ LP      & 0.0899 & 0.0593  &  0.0549  & 0.0368  \\
\bottomrule
\end{tabular}
\end{sc}
\end{small}
\end{center}
\vskip -0.1in
\end{table}

\begin{table}[H]
\caption{ 6-DOF K Step Mean Abs Prediction Error and Explained Variance}
\label{tab:6-DOF_perf}
\vskip 0.15in
\begin{center}
\begin{small}
\begin{sc}
\begin{tabular}{lccccc}
\toprule
\midrule
K       & 1 & 10 & 30 & 60\\
GT Pred     & 0.0101 &  0.0089 & 0.0090   &  0.0085  \\
GT EV       & 0.9845 &  0.9739  & 0.9562  &  0.8633\\
DTM Pred    & 0.0094 & 0.0129  & 0.0120  & 0.0093   \\
DTM EV      & 0.9225 & 0.9442  &  0.8633  & 0.5988   \\
\bottomrule
\end{tabular}
\end{sc}
\end{small}
\end{center}
\vskip -0.1in
\end{table}

\subsection{Accelerated RL using a Predictive Model}
\label{section:Dyna}
In this experiment we demonstrate an improvement in sample effiency obtained using PPO in conjunction with our model. The model $\mu$, policy $\pi$, and value function $\nu$ learn concurrently in the 3-DOF Mars environment.  After each update of $\mu$, $\pi$, $\nu$ from rollouts $\mathcal{R}$, $s$ observations are sampled from $\mathcal{R}$.  These are fed through $\pi$ and $\mu$ in parallel for $k$ steps to create a set $\mathcal{S}$ of simulated rollouts as shown in Algorithm \ref{alg:dyna}. Note that these segments do not result in a full episode, and consequently, when we update the policy using the simulated rollouts, we need to take care discounting the simulated rewards to generate the empirical return for updating the policy. Concretely, we need to use the correct $n$-step temporal difference return, and use this to compute the advantages $\mathcal{A}$; this is implemented in lines 12 through 17 of the algorithm.

\begin{figure}[!htbp]
\vskip 0.2in
\begin{center}
\centerline{\includegraphics[width=\columnwidth]{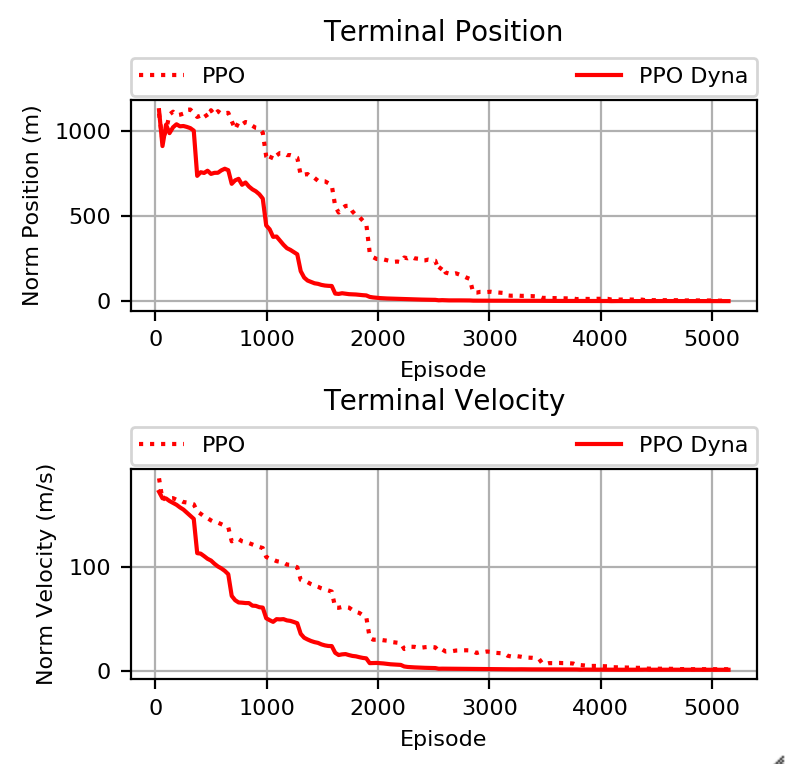}}
\caption{Dyna versus Standard PPO Performance}
\label{fig:Dyna_compare}
\end{center}
\vskip -0.2in
\end{figure}

Figure \ref{fig:Dyna_compare} plots the norm of the lander terminal position and velocity as a function of training episode for a policy optimized using PPO and PPO with Dyna over the first 5000 episodes.  The simulated rollouts were generated using 20,000 observations sampled from the previous rollouts with k=10, resulting in simulated rollouts containing 200,000 tuples of observations, actions, and advantages. We find that PPO enhanced with Dyna converges roughly twice as fast. 

\begin{algorithm}[tb]
   \caption{Dyna Algorithm}
   \label{alg:dyna}
\begin{algorithmic}[1]
    \STATE Initialize Model, Policy, Value Function $\mathcal{M}, \mathcal{P}, \mathcal{V} $
    \FOR{ update = 1 , m} 
        \STATE Collect rollouts $\mathcal{R}$ from running E episodes
        \STATE Update $\mu,\pi,\nu$ using $\mathcal{R}$ 
        \STATE Randomly select the set $\mathcal{O}$ of observations from $\mathcal{R}$
        \STATE $\mathcal{S} = \{\}$
        \FOR{t=0,k-1} 
            \STATE $\mathbf{A}_{t}=\pi(\mathbf{O_t})$
            \STATE $\mathbf{O_{t+1},R_{t+1}}=\mu(\mathbf{A_{t}},\mathbf{0})$
            \STATE append $\mathbf{O}_t,\mathbf{A}_t,\mathbf{R}_{t+1}$ to $\mathcal{S}$
        \ENDFOR
        \STATE $\mathbf{R}_f= \nu(\mathbf{O}_{t+1})$ 
        \FOR{t=0,k-1} 
            \STATE $\mathbf{E}_{t}=\sum_{\tau=t}^{k}\gamma^{\tau-t}\mathbf{R}_{\tau}$ 
            \STATE $\mathbf{E}_t=\mathbf{E}_t+\gamma^{K-t}\mathbf{R}_f$ 
            \STATE Add $\mathcal{A}$ = $\mathbf{E}_t - \nu(\mathbf{O}_t)$ to $\mathcal{S}$
        \ENDFOR
        \STATE Update $\pi$ using $\mathcal{S}$ 
    \ENDFOR
\end{algorithmic}
\end{algorithm}

\section{Discussion}

We have presented a novel predictive coding model architecture capable of generating accurate trajectory predictions over long horizons. Through a series of experiments, we have shown that our enhancements to  predictive coding outperforms a standard implementation for long-horizon predictions, and performs well predicting both the agent's ground truth state and simulated altimeter readings open loop over an entire episode during a 6-DOF Mars powered descent phase. The ability to generate long horizon open loop trajectory predictions is extremely useful for both model based reinforcement learning and model predictive control.  We demonstrated the ability of the model to reduce the sample complexity of proximal policy optimization for a Mars 3-DOF powered descent phase task.  Future work will extend the model to predicting observations in higher dimensional spaces consistent with flash LIDAR and electro-optical sensors.

\bibliography{PCM}

\begin{thebibliography}{16}
\providecommand{\natexlab}[1]{#1}
\providecommand{\url}[1]{\texttt{#1}}
\expandafter\ifx\csname urlstyle\endcsname\relax
  \providecommand{\doi}[1]{doi: #1}\else
  \providecommand{\doi}{doi: \begingroup \urlstyle{rm}\Url}\fi

\bibitem[Bengio et~al.(2009)Bengio, Louradour, Collobert, and
  Weston]{bengio2009curriculum}
Bengio, Y., Louradour, J., Collobert, R., and Weston, J.
\newblock Curriculum learning.
\newblock In \emph{Proceedings of the 26th annual international conference on
  machine learning}, pp.\  41--48. ACM, 2009.

\bibitem[Chung et~al.(2015)Chung, Gulcehre, Cho, and Bengio]{chung2015gated}
Chung, J., Gulcehre, C., Cho, K., and Bengio, Y.
\newblock Gated feedback recurrent neural networks.
\newblock In \emph{International Conference on Machine Learning}, pp.\
  2067--2075, 2015.

\bibitem[D'Souza \& D'Souza(1997)D'Souza and D'Souza]{d1997optimal}
D'Souza, C. and D'Souza, C.
\newblock An optimal guidance law for planetary landing.
\newblock In \emph{Guidance, Navigation, and Control Conference}, pp.\  3709,
  1997.

\bibitem[Feinberg et~al.(2018)Feinberg, Wan, Stoica, Jordan, Gonzalez, and
  Levine]{feinberg2018model}
Feinberg, V., Wan, A., Stoica, I., Jordan, M.~I., Gonzalez, J.~E., and Levine,
  S.
\newblock Model-based value estimation for efficient model-free reinforcement
  learning.
\newblock \emph{arXiv preprint arXiv:1803.00101}, 2018.

\bibitem[Finn \& Levine(2017)Finn and Levine]{finn2017deep}
Finn, C. and Levine, S.
\newblock Deep visual foresight for planning robot motion.
\newblock In \emph{Robotics and Automation (ICRA), 2017 IEEE International
  Conference on}, pp.\  2786--2793. IEEE, 2017.

\bibitem[Gaudet et~al.(2018)Gaudet, Linares, and Furfaro]{gaudet2018deep}
Gaudet, B., Linares, R., and Furfaro, R.
\newblock Deep reinforcement learning for six degree-of-freedom planetary
  powered descent and landing.
\newblock \emph{arXiv preprint arXiv:1810.08719}, 2018.

\bibitem[Gu et~al.(2016)Gu, Lillicrap, Sutskever, and Levine]{gu2016continuous}
Gu, S., Lillicrap, T., Sutskever, I., and Levine, S.
\newblock Continuous deep q-learning with model-based acceleration.
\newblock In \emph{International Conference on Machine Learning}, pp.\
  2829--2838, 2016.

\bibitem[Lillicrap et~al.(2015)Lillicrap, Hunt, Pritzel, Heess, Erez, Tassa,
  Silver, and Wierstra]{lillicrap2015continuous}
Lillicrap, T.~P., Hunt, J.~J., Pritzel, A., Heess, N., Erez, T., Tassa, Y.,
  Silver, D., and Wierstra, D.
\newblock Continuous control with deep reinforcement learning.
\newblock \emph{arXiv preprint arXiv:1509.02971}, 2015.

\bibitem[Lotter et~al.(2016)Lotter, Kreiman, and Cox]{lotter2016deep}
Lotter, W., Kreiman, G., and Cox, D.
\newblock Deep predictive coding networks for video prediction and unsupervised
  learning.
\newblock \emph{arXiv preprint arXiv:1605.08104}, 2016.

\bibitem[Nagabandi et~al.(2018)Nagabandi, Kahn, Fearing, and
  Levine]{nagabandi2018neural}
Nagabandi, A., Kahn, G., Fearing, R.~S., and Levine, S.
\newblock Neural network dynamics for model-based deep reinforcement learning
  with model-free fine-tuning.
\newblock In \emph{2018 IEEE International Conference on Robotics and
  Automation (ICRA)}, pp.\  7559--7566. IEEE, 2018.

\bibitem[Oh et~al.(2015)Oh, Guo, Lee, Lewis, and Singh]{oh2015action}
Oh, J., Guo, X., Lee, H., Lewis, R.~L., and Singh, S.
\newblock Action-conditional video prediction using deep networks in atari
  games.
\newblock In \emph{Advances in neural information processing systems}, pp.\
  2863--2871, 2015.

\bibitem[Rao \& Ballard(1999)Rao and Ballard]{rao1999predictive}
Rao, R.~P. and Ballard, D.~H.
\newblock Predictive coding in the visual cortex: a functional interpretation
  of some extra-classical receptive-field effects.
\newblock \emph{Nature neuroscience}, 2\penalty0 (1):\penalty0 79, 1999.

\bibitem[Schulman et~al.(2015)Schulman, Levine, Abbeel, Jordan, and
  Moritz]{schulman2015trust}
Schulman, J., Levine, S., Abbeel, P., Jordan, M., and Moritz, P.
\newblock Trust region policy optimization.
\newblock In \emph{International Conference on Machine Learning}, pp.\
  1889--1897, 2015.

\bibitem[Schulman et~al.(2017)Schulman, Wolski, Dhariwal, Radford, and
  Klimov]{schulman2017proximal}
Schulman, J., Wolski, F., Dhariwal, P., Radford, A., and Klimov, O.
\newblock Proximal policy optimization algorithms.
\newblock \emph{arXiv preprint arXiv:1707.06347}, 2017.

\bibitem[Sutton(1990)]{sutton1990integrated}
Sutton, R.
\newblock Integrated architectures for learning, planning, and reacting based
  on approximating integrated architectures for learning, planning, and
  reacting based on approximating dynamic programming.
\newblock In \emph{Proceedings of the International Machine Learning
  Conference}, pp.\  212--218, 1990.

\bibitem[Wang et~al.(2017)Wang, Long, Wang, Gao, and Philip]{wang2017predrnn}
Wang, Y., Long, M., Wang, J., Gao, Z., and Philip, S.~Y.
\newblock Predrnn: Recurrent neural networks for predictive learning using
  spatiotemporal lstms.
\newblock In \emph{Advances in Neural Information Processing Systems}, pp.\
  879--888, 2017.

\end{thebibliography}
\bibliographystyle{icml2019}

\end{document}